\documentclass[10pt,twocolumn,letterpaper]{article}

\usepackage{cvpr}              
\usepackage{amsmath, amsthm} 
\newtheorem{theorem}{Theorem}[section] 
\newtheorem{corollary}[theorem]{Corollary}
\usepackage{bbding}
\usepackage{algorithmic}
\usepackage{algorithm}
\usepackage{listings}
\usepackage{xcolor}
\definecolor{cvprblue}{rgb}{0.21,0.49,0.74}
\usepackage[pagebackref,breaklinks,colorlinks,allcolors=cvprblue]{hyperref}
\usepackage{multirow} 
\def\paperID{} 
\def\confName{CVPR}
\def\confYear{2026}

\title{CAR-SAM: Cross-Attention Reconstruction for Post-Training Quantization of the Segment Anything Model}

\author{Houji Wen$^{1,2 \ast \ddagger}$, ~~
    Jiangyong Yu$^{2 \ast}$, ~~
    Jun Li$^{1 \dagger}$, ~~
      Dawei Yang$^{2  \dagger}$\\
    $^1$Nanjing University of Science and Technology 
    ~~~$^2$Houmo AI \\
    {\tt\small dawei.yang@houmo.ai junli@njust.edu.cn}
}

\begin{document}
\maketitle
\renewcommand{\thefootnote}{}

\begin{abstract}
Segment Anything Models (SAMs) are extensively used in computer vision for universal image segmentation, but deploying them on resource-constrained devices is challenging due to their high computational and memory demands. Post-Training Quantization (PTQ) is a widely used technique for model compression and acceleration. However, existing PTQ methods fail to consider the cross-attention architecture in the SAM decoder.
This degradation primarily stems from the unique challenges posed by SAMs: 
(1) Attention dissipation, where the attention information in the decoder, which is crucial for representing segmentation masks, collapses into a diffuse and non-semantic form under low-bit quantization;
and (2) Reconstruction oscillation, where bidirectional coupling within the two-way transformer introduces cross-branch error interference and destabilizes convergence. 
To tackle these issues, we propose CAR-SAM, a unified quantization framework tailored for SAMs. 
Firstly, to mitigate attention dissipation, we introduce MatMul-Aware Compensation (MAC) mechanism that transfers activation-induced quantization errors from MatMul to preceding linear weights.
Secondly, to mitigate oscillation in decoder optimization, we develop a Joint Cross-Attention Reconstruction (JCAR) strategy that jointly reconstructs coupled attention branches, suppressing oscillatory behavior and promoting stable convergence.
Extensive experiments show that CAR-SAM robustly quantizes SAM models down to 4-bit precision, surpassing existing methods by 14.6\% and 6.6\% mAP on SAM-B and SAM-L respectively.
\end{abstract} 
\footnote{${\ast}$  Equal contribution, $\dagger$ Corresponding author, $\ddagger$ This work was conducted during his internship at Houmo.}
\renewcommand{\thefootnote}{\arabic{footnote}}
\addtocounter{footnote}{-2}

\section{Introduction}
\label{sec:intro}
Segment Anything Model \cite{kirillov2023segment} (SAM) and its successor, SAM2 \cite{ravi2024sam}, represent a significant breakthrough in computer vision. Their ability to perform prompt-guided, zero-shot segmentation on arbitrary objects has unlocked new capabilities across a vast range of applications. Despite SAM’s strong segmentation capability, the Large and Huge variants exceed 1 B parameters and 1 T FLOPs, posing serious challenges for edge deployment.

\begin{figure}[t]
\centering
\subcaptionbox{QDrop\label{fig:intro_a}}[.24\linewidth]{%
    \includegraphics[width=\linewidth]{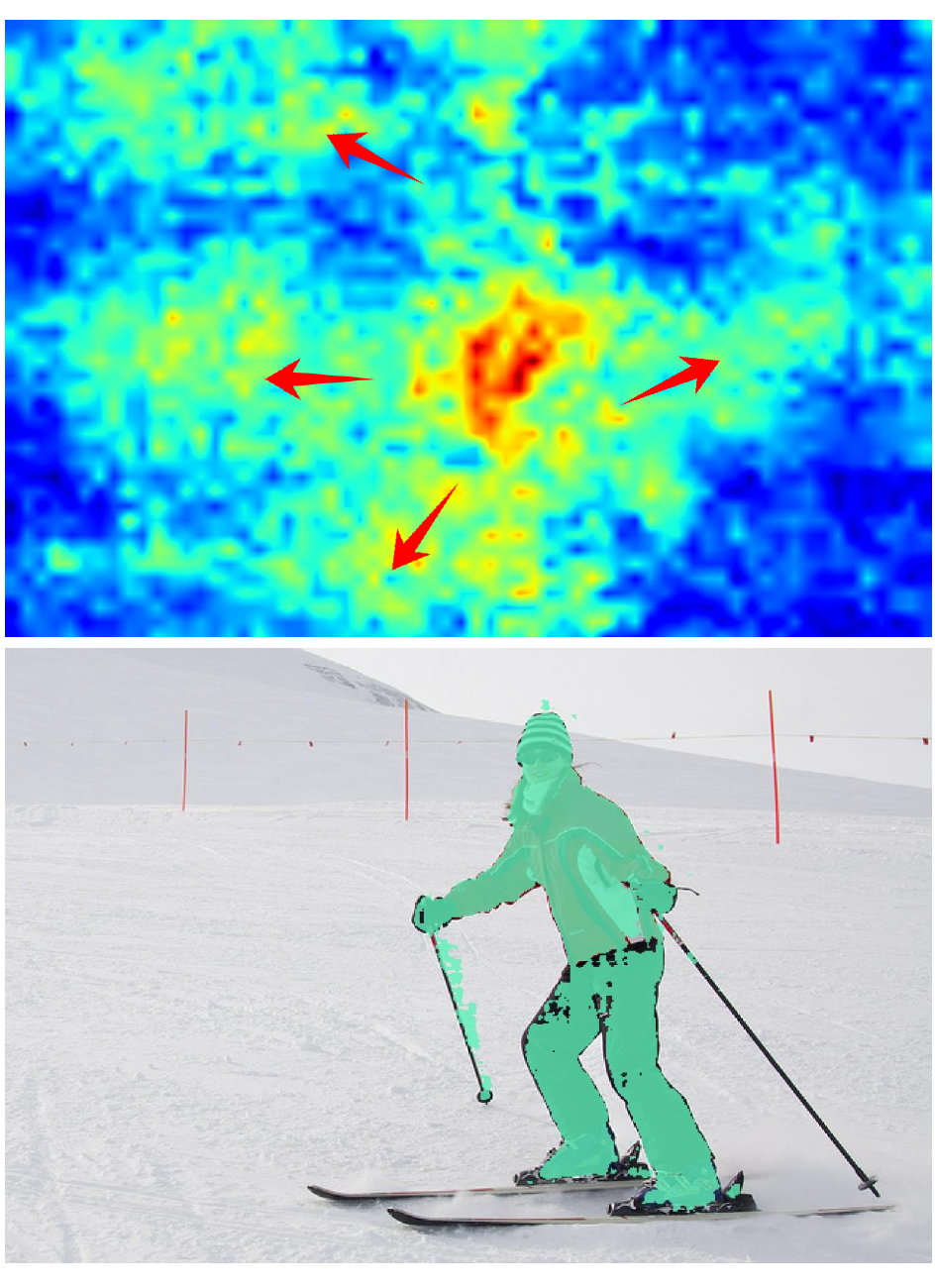}}
\subcaptionbox{PTQ4SAM\label{fig:intro_b}}[.24\linewidth]{%
    \includegraphics[width=\linewidth]{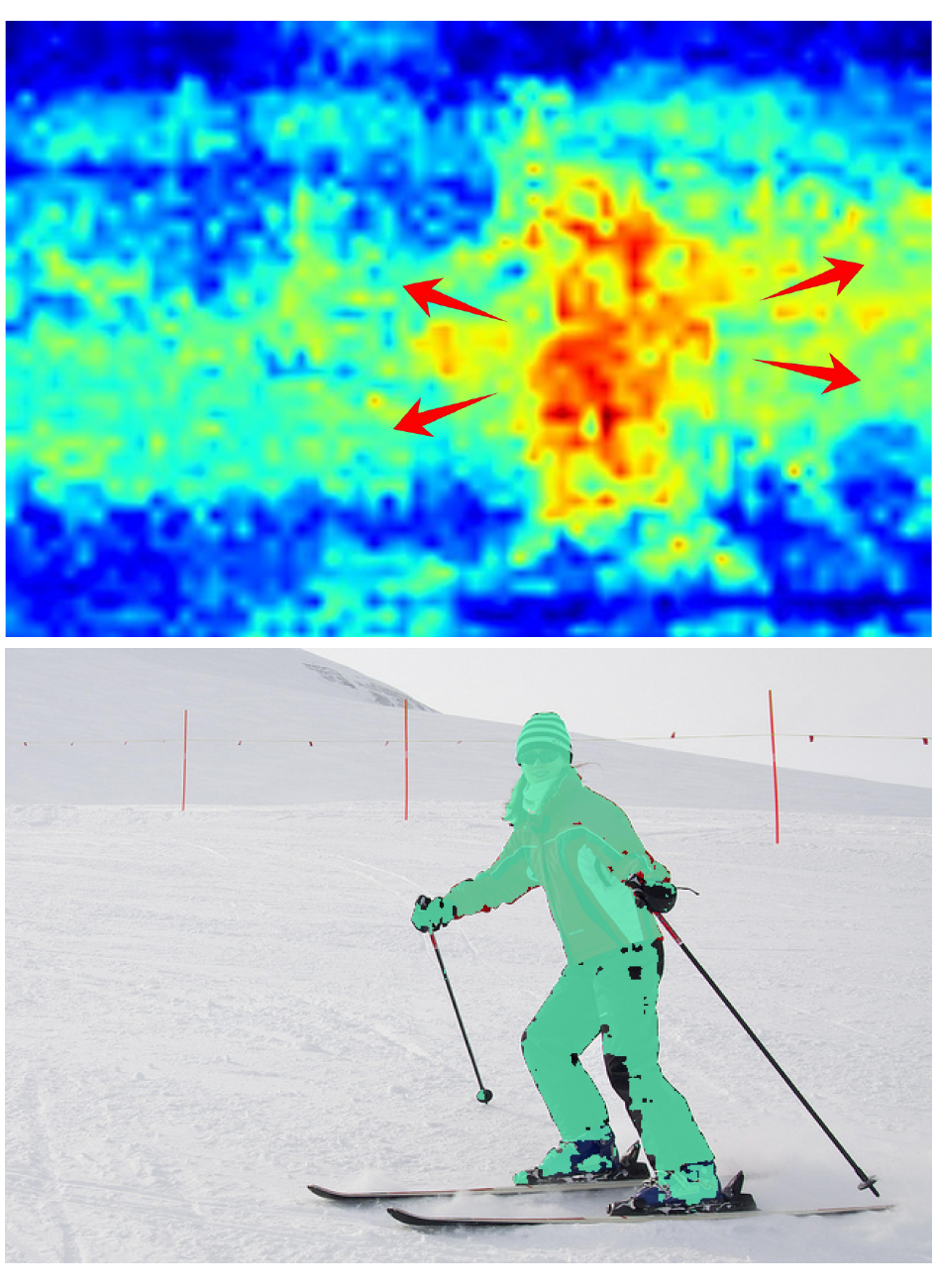}}
\subcaptionbox{FP\label{fig:intro_c}}[.24\linewidth]{%
    \includegraphics[width=\linewidth]{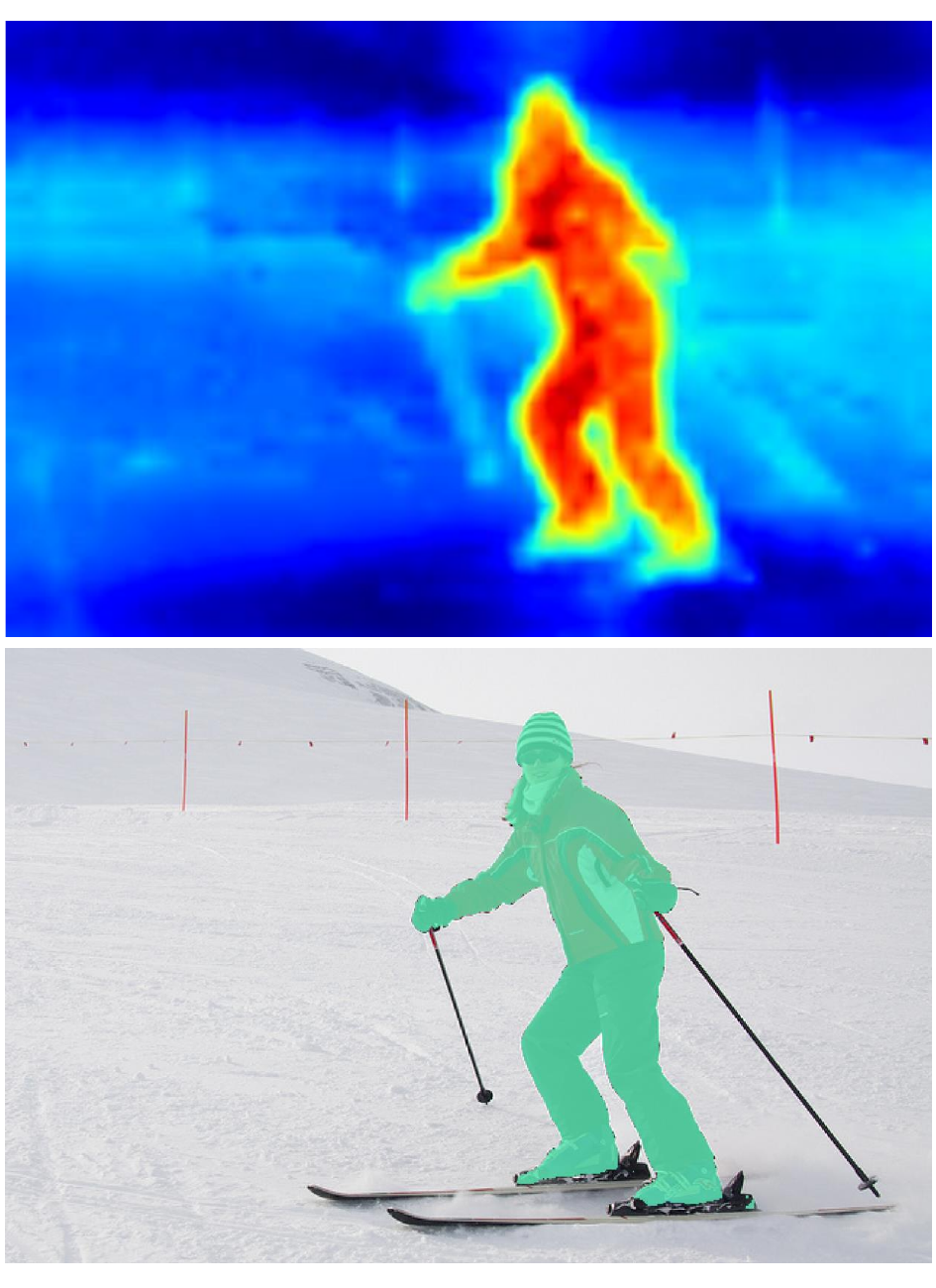}}
\subcaptionbox{Ours\label{fig:intro_d}}[.24\linewidth]{%
    \includegraphics[width=\linewidth]{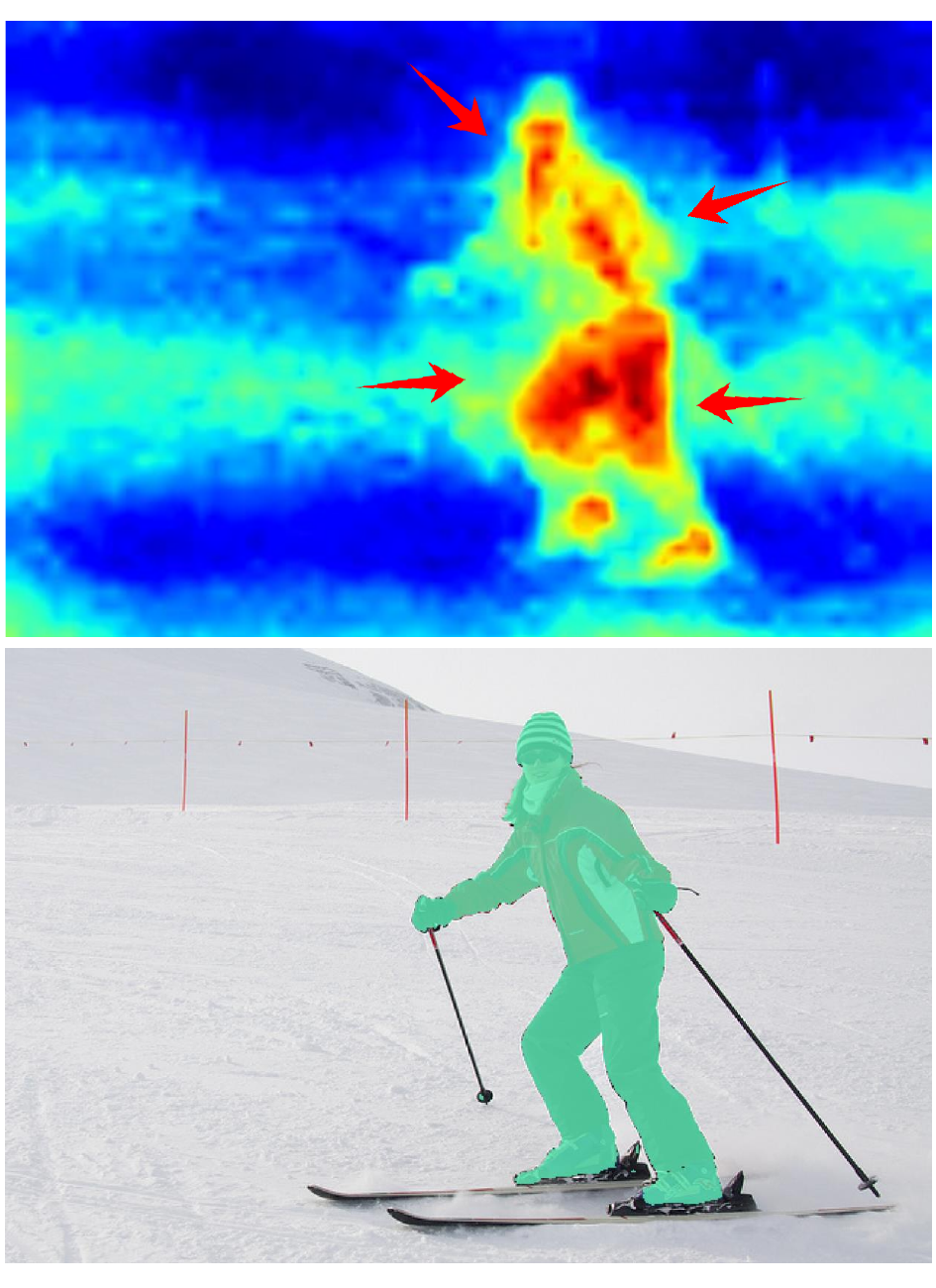}}

\caption{4-bit quantized segmentation results using different methods based on SAM-B. Each example shows the attention heatmap of mask token (top) and corresponding mask prediction (bottom). (a) QDrop and (b) PTQ4SAM suffers from attention dissipation and poor segmentation. (c) Full-precision baseline. (d) Our method preserves attention focus and yields accurate masks.}
\label{fig1}
\end{figure}

Post-training quantization (PTQ) offers a lightweight way to compress neural networks without retraining. 
Existing PTQ methods for ViTs, such as PTQ4ViT \cite{yuan2022ptq4vit}, BRECQ \cite{li2021brecq}, QDrop \cite{wei2022qdrop}, and APHQ-ViT \cite{wu2025aphq}, have explored various reconstruction-based strategy; however, these approaches primarily target encoder-only architectures. Recent attempts PTQ4SAM \cite{lv2024ptq4sam}, MIX-QSAM \cite{ranjan2025mix}, Pq-sam \cite{liu2024pq} provide partial solutions for quantizing SAM, but they still fail to address the fundamental challenge posed by the decoder’s bidirectional cross-attention structure. As a result, SAM still suffers from a noticeable mAP drop under W4A4 quantization. To better understand why SAM suffers from such severe degradation, we therefore investigate what fundamentally causes this instability and summarize our findings into two main challenges.

1) \textbf{Attention Dissipation} in the decoder’s image-to-token cross-attention. As shown in Fig. \ref{fig1}, low-bit quantization causes the attention distribution to collapse into diffuse regions, weakening image–prompt interaction and degrading segmentation masks. This degradation arises because the attention score is computed by a matrix multiplication (MatMul) between heterogeneous modalities, where queries from prompt and keys from image. The two modalities show distinct statistics: image embeddings are narrowly distributed within [-14, 15], while prompt tokens span a much wider range of [-40, 65]. Quantizing this matmul inputs amplifies scale mismatch, distorts the score distribution, and flattens attention. Existing PTQ methods mainly compensate linear layers, but do not compensate the matmul itself where inter-modal statistics interact, leaving the dissipation unaddressed 
\begin{figure}[t]
\centering
\includegraphics[width=1.0\linewidth]{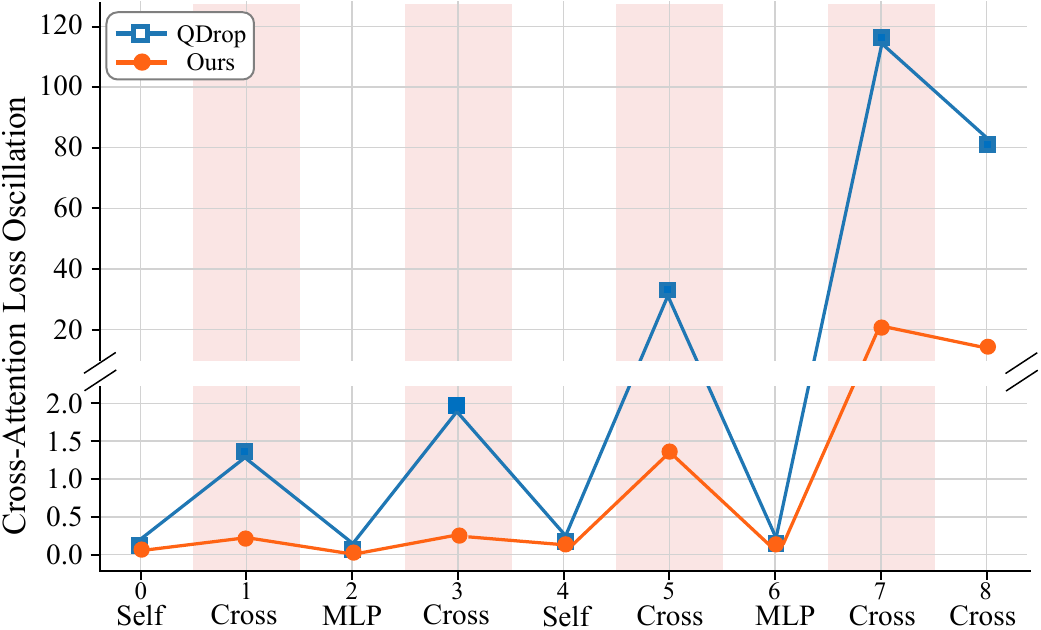}
\caption{Reconstruction loss across decoder blocks in SAM. Reconstruction loss of different module types (Self-Attention, Cross-Attention, and MLP) under 4-bit quantization. The loss exhibits strong oscillations along the decoding depth, with the largest fluctuations appearing at cross-attention layers, indicating the high quantization sensitivity of the bidirectionally coupled cross-attention modules.}
\label{fig:fig2}
\end{figure}

2) \textbf{Reconstruction Oscillation} in the decoder. A second obstacle arises during the reconstruction of the decoder, where we observe that the loss curve oscillates strongly rather than converging smoothly, as illustrated in Fig.~\ref{fig:fig2}. Consequently, the quantized model fails to reach a global optimum and instead oscillates around local stationary points. 
This behavior differs markedly from that of CNNs or ViTs, which follow sequential processing pipelines. 
In contrast, the cross-attention modules in the decoder employ partially parallel pipelines that process the image and the prompt, which continuously exchange information through bidirectional cross-attention. 
Errors introduced in one pipeline can instantly affect the other through cross-attention, creating cross-modal feedback loops that cause the loss to oscillate. 

To address the above challenges, we propose a post-training quantization framework called CAR-SAM, specifically designed for SAM and SAM2.
Specifically, to mitigate the problem of \textit{attention dissipation}, we propose an \textbf{MatMul-Aware Compensation (MAC)} method, which extends linear-layer compensation into a matmul-aware formulation, that explicitly considers the disparity between two modalities. Specifically, this method transfers the quantization error introduced by the activation quantization of matmul inputs back to the preceding linear layer’s weights, thereby achieving a form of cross-layer compensation. 
To stabilize the \textit{Optimization Oscillation}, we further design \textbf{Joint Cross-Attention Reconstruction (JCAR)} strategy for the coupled cross-attention modules. Our mathematical analysis demonstrates that the loss term of this architecture contains not only upstream error-propagation components but also cross-branch coupling terms arising from bidirectional interactions. We prove that optimizing these coupled modules jointly at the cross-attention level yields the most stable and accurate reconstruction.

Our main contributions are summarized as follows:
\begin{itemize}
    \item We present CAR-SAM, the first unified post-training quantization framework compatible with both SAM and SAM2, tailored to their distinct architecture designs.
    \item We propose a MatMul-Aware Compensation (MAC) mechanism that transfers quantization errors from matmul activations to preceding linear weights, achieving cross-layer compensation and mitigating attention dissipation.
    \item We introduce a Joint Cross-Attention Reconstruction (JCAR) strategy to stabilize optimization oscillation in coupled cross-attention modules. Our analysis shows that joint optimization effectively suppresses cross-branch error coupling.
    \item Extensive experiments show that CAR-SAM consistently outperforms prior methods across various zero-shot segmentation tasks, delivering robust 4-bit performance.
\end{itemize}
\section{Related Work}
\label{sec:Related_Work}
\subsection{Segment Anything Models}
Meta AI’s Segment Anything Model (SAM) introduced prompt-driven, zero-shot image segmentation and has since been verified across several domains. Its successor, SAM2, not only improves segmentation accuracy but also extends the model’s capabilities to the video domain, enabling effective video segmentation. SAM family models have demonstrated broad applicability across a wide range of downstream tasks ~\cite{zhu2024medical,ren2024grounded,yang2023sam3d,williams2024leaf,zhou2023dsec}.
Despite the model’s broad adoption, its substantial computational cost remains a major limitation and has prompted extensive work on more efficient SAM variants. One popular strategy for compressing SAM is knowledge distillation. MobileSAM, Tiny-SAM, and EdgeSAM~\citep{zhang2023faster,shu2025tinysam,zhou2023edgesam} replace the heavy ViT encoder with lightweight CNNs or compact ViTs. Alternatively, SlimSAM~\citep{chen2024slimsam} adopts an iterative pruning-distillation framework to progressively compress decoupled substructures.
\subsection{Post-Training Quantization for SAM}
PTQ directly quantizes a pre-trained model only with a small, unlabeled calibration set, making it a fast and efficient solution. PTQ methods can be broadly categorized into statistical-based and learning-based approaches. Statistical PTQ methods ~\cite{yuan2022ptq4vit,lin2021fq}  focus on minimizing quantization error by analyzing and adapting to the statistical properties of network parameters. Learning-based PTQ \cite{esser2019learned, bhalgat2020lsq+}extends statistical methods by employing optimization or reconstruction techniques to further reduce quantization error. These methods typically calibrate a quantized model layer-by-layer or block-by-block, minimizing the reconstruction loss between the quantized and full-precision outputs. AdaRound~\cite{nagel2020up} was a pioneering work that optimized the weight rounding operation to minimize overall model loss. Building on this, BRECQ~\cite{li2021brecq} introduced a more efficient block-wise reconstruction algorithm, while QDrop~\cite{wei2022qdrop} incorporated dropout during reconstruction to improve the robustness of the quantized model.  

Low-bit quantization of SAM has been explored in several studies, among which three methods are most representative. PQ-SAM mitigates encoder outliers using Grouped Activation Distribution Transformation (GADT), but our experiments show that the decoder, rather than the encoder, is the main source of degradation under low-bit quantization. PTQ4SAM models bimodal activations with a Bimodal Integration scheme and applies logarithmic quantization to softmax outputs, although this approach lacks generality because SAM2 no longer exhibits such bimodal behavior. Mix-QSAM adopts a mixed-precision strategy based on KL-guided layer importance and solves an integer programming problem for bit allocation, but its high computational cost and unverified performance on SAM2 limit its practicality.

\section{Method}

\subsection{Preliminaries}
For a $k$-bit uniform quantization scheme, the quantization and dequantization processes can be formally defined as
\begin{gather}
    x_{q} = \text{clamp}( \lfloor \frac{x}{s} \rceil+z , 0, 2^k-1),\\
    \hat{x} = s \cdot (x_{q} - z) \approx x,
\end{gather}
where $s$ and $z$ denote the scaling factor and zero point, respectively, and $\lfloor \cdot \rceil$ represents the round-to-nearest operator. $x$ and $\hat{x}$ are the original floating-point and dequantized values, respectively, while $x_q$ is the quantized integer representation. 

PTQ commonly adopts reconstruction-based optimization to learn quantization parameters that minimize the discrepancy between quantized and full-precision outputs. 
\[
\min_{\mathbf{s},\,\boldsymbol{\alpha}}
\;\bigl\|
F(\hat{\mathbf{x}}, \hat{\mathbf{w}}) - F(\mathbf{x}, \mathbf{w})
\bigr\|_2^2,
\]
where $\mathbf{s}$ and $\boldsymbol{\alpha}$ denote the activation scaling factor and weight rounding offset respectively. 
$\mathbf{x}$ and $\mathbf{w}$ are the original floating-point activation and weight, while $\hat{\mathbf{x}}$ and $\hat{\mathbf{w}}$ are their dequantized counterparts.  
$F(\cdot)$ represents the attention block being reconstructed.  
The objective ensures that the quantized output $F(\hat{\mathbf{x}}, \hat{\mathbf{w}})$ remains well-aligned with the full-precision reference $F(\mathbf{x}, \mathbf{w})$ by jointly optimizing both quantization parameters.

\begin{figure*}[!ht]
    \centering
    \includegraphics[width=1\linewidth]{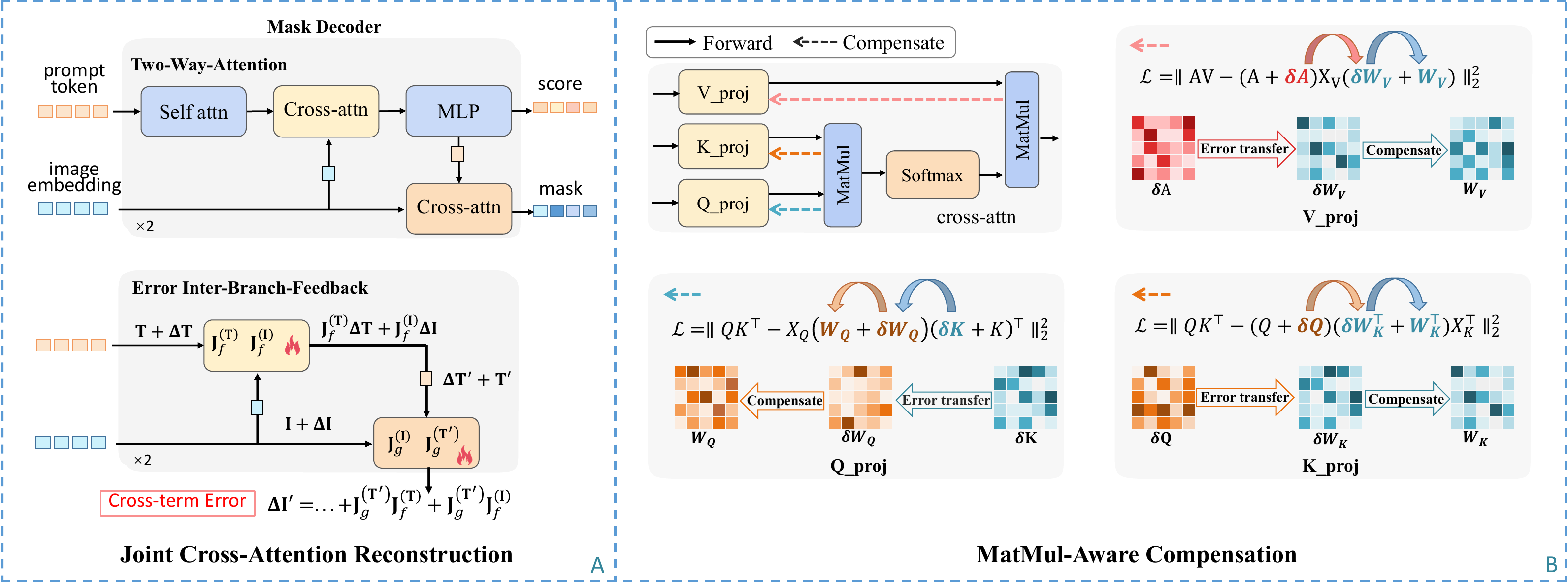}
    \caption{Overview of the proposed framework. 
        The \textbf{top-left} panel illustrates the overall architecture of the SAM mask decoder. 
        The \textbf{bottom-left} diagram depicts the error propagation between its two cross-attention modules: 
        when reconstructing the latter cross-attention, the Jacobian term $J_f$ from the preceding module introduces a cross-term error that couples their optimization dynamics. 
        The \textbf{right} panel shows our \textbf{MatMul-Aware Compensation (MAC)} mechanism. 
        For both matmul operations ($QK^{\top}$ and $AV$), we estimate the quantization errors of the input variables $Q$, $K$, and $V$, 
        and propagate these errors backward as compensation into the preceding projection layers, effectively achieving cross-layer error compensation.
        }
            \label{fig:frame}
    \vspace{-1mm}
\end{figure*}

\subsection{MatMul Aware Compensation}
Quantizing activations remains one of the most challenging aspects of PTQ, since activations vary dynamically with the input distribution. Early methods such as GPTQ~\cite{frantar2022gptq}, BRECQ~\cite{li2021brecq}, and AdaRound~\cite{nagel2020up} focus solely on weight quantization, reconstructing layer outputs to minimize functional errors after quantization.
Later extensions, including GPTAQ~\cite{li2025gptaq}, ERQ~\cite{zhong2024erq}, and QDrop~\cite{wei2022qdrop}, further consider activation quantization by optimizing weights to compensate for activation-induced errors, achieving better alignment between quantized and full-precision outputs.

However, the attention mechanism involves two linear operations: the projection layers (Q, K, V) and the matrix multiplication (MatMul) for attention-score computation. Most existing methods, such as ERQ~\cite{zhong2024erq}, QwT~\cite{xu2010qwt}, and SVDQuant~\cite{li2024svdquant}, focus solely on compensating the projection linears, leaving the MatMul operation unmodeled. This omission becomes critical in SAM’s decoder, thereby causing attention dissipation. 

Motivated by this limitation, we formulate explicit reconstruction objectives for the $QK^{\top}$ and $AV$ MatMul operations, allowing us to compensate for activation quantization errors introduced by the activation quantization of MatMul inputs. 
To compensate for the quantization error introduced by the MatMul operation in attention computation, we formulate the following reconstruction objective:
\begin{equation}
\mathcal{L}^{\text{mse}} = \mathbb{E}\left[ \left\| \mathbf{Q}\mathbf{K}^{\top} - \hat{\mathbf{Q}}\hat{\mathbf{K}}^{\top} \right\|_2^2 \right]
    \label{eq:qk_loss}
\end{equation}
where $\mathbf{\hat{Q}}$ and $\hat{\mathbf{K}}^{\top}$ denote quantized queries and keys respectively, $\mathbf{Q}$ and $\mathbf{K}^{\top}$ denote full-precision (FP) queries and keys respectively. Our goal is to minimize the reconstruction discrepancy between the full-precision and quantized MatMul outputs.

\paragraph{Propagation to $Q_{\text{proj}}$.}
To model the quantization-induced perturbations explicitly, we parameterize:
\begin{equation}
    \hat{\mathbf{Q}} = \mathbf{X}_Q (\mathbf{W}_Q + \delta \mathbf{W}_Q), \quad
    \hat{\mathbf{K}} = \mathbf{K} + \delta \mathbf{K}.
\end{equation}
where $\mathbf{X}_Q$ is the fp input of  $\mathbf{Q_{proj}}$, $\delta \mathbf{W}_Q$ denotes the compensation term, and $\hat{\mathbf{K}} = \mathbf{K} + \delta \mathbf{K}$ is the error propagated from the input quantization error of the MatMul operands. 
Substituting these into Eq.~(\ref{eq:qk_loss}) yields:
\begin{equation}
    \begin{split}
\mathcal{L}^{\text{mse}} &= \bigl\| \mathbf{Q}\mathbf{K}^{\top} - \mathbf{X}_Q (\mathbf{W}_Q + \delta \mathbf{W}_Q)(\mathbf{K} + \delta \mathbf{K})^{\top} \bigr\|_2^2 \\
&\quad + \lambda \| \delta \mathbf{W}_Q \|_2^2.
    \end{split}
    \label{eq:k_loss_expand}
\end{equation}
where the last term introduces an $L_2$ regularization to prevent the compensation term $\delta \mathbf{W}_Q $ from overfitting or dominating the original weight $ \mathbf{W}_Q $. The selection of the regularization coefficient $\lambda$ is obtained through the SVD decomposition of $\mathbf{X_Q}^{\top} \mathbf{X_Q}$, with details derivation provided in Appendix A.

By expanding Eq.~(\ref{eq:k_loss_expand}) and taking the derivative of $\mathcal{L}$ with respect to $\mathbf{W}_Q$ and setting it to zero, we rearrange the terms, we obtain a Sylvester-type matrix equation:
\begin{equation}
    \mathbf{A} \mathbf{X} + \mathbf{X} \mathbf{B} = \mathbf{C}.
    \label{eq:Sylvester_eq}
\end{equation}
where
\begin{equation}
    \begin{aligned}
\mathbf{A} &= (\mathbf{X}_Q^{\top} \mathbf{X}_Q)^{-1}\lambda \mathbf{I}, & \mathbf{B} &= \hat{\mathbf{K}}^{\top}\hat{\mathbf{K}}, \\
\mathbf{C} &= \mathbf{W}_Q \,\delta \mathbf{K}^{\top}\hat{\mathbf{K}},   & \mathbf{X} &= \delta \mathbf{W}_Q.
    \end{aligned}
    \label{eq:delta Wq solution}
\end{equation}
This equation takes the form of a Sylvester-type matrix equation, which can be efficiently solved either in closed form or through iterative methods such as the Bartels–Stewart algorithm.
Once the solution $\mathbf{X}$ is obtained, the compensation term $\boldsymbol{\delta}\mathbf{W}_{\mathbf{Q}}$ is derived and merged into the original weight as  $\mathbf{W_Q} \leftarrow \mathbf{W_Q} + \delta \mathbf{W}_Q$. 
This formulation reveals an instructive insight: the quantization error of $\mathbf{K}$ does not have to be locally compensated on the key branch; it can be re-channeled into the upstream $\mathbf{Q}_{\text{proj}}$, allowing the matmul output to retain its alignment without explicitly modifying the quantized $\mathbf{K}$.

\paragraph{Propagation to $K_{\text{proj}}$.}
By taking similar steps, to propagate the quantization error of $\boldsymbol{\delta}\mathbf{Q}$ into the key projection $\mathbf{W_k}$, 
we reformulate Eq.~(\ref{eq:qk_loss}) by defining:
\begin{equation}
    \hat{\mathbf{K}} = \mathbf{X}_{\mathbf{K}} (\mathbf{W}_{\mathbf{K}} + \boldsymbol{\delta}\mathbf{W}_{\mathbf{K}}), \quad
    \hat{\mathbf{Q}} = \mathbf{Q} + \boldsymbol{\delta}\mathbf{Q}.
\end{equation}
The reconstruction objective is then expressed as:
\begin{equation}
    \begin{split}
    \mathcal{L}^{\text{mse}} &= \bigl\| \mathbf{Q}\mathbf{K}^{\top} - (\mathbf{Q} + \boldsymbol{\delta}\mathbf{Q})(\mathbf{W}_{\mathbf{K}}^{\top} + \boldsymbol{\delta}\mathbf{W}_{\mathbf{K}}^{\top})\mathbf{X}_{\mathbf{K}}^{\top} \bigr\|_2^2 \\
    &\quad + \lambda \|\boldsymbol{\delta}\mathbf{W}_{\mathbf{K}}\|_2^2.
    \end{split}
\end{equation}
Following the same derivation procedure, 
the first-order condition with respect to $\boldsymbol{\delta}\mathbf{W}_{\mathbf{K}}$ yields the same Sylvester-type Eq.~(\ref{eq:Sylvester_eq}), where
\begin{equation}
    \begin{aligned}
        \mathbf{A} &= (\mathbf{X}_{\mathbf{K}}^{\top} \mathbf{X}_{\mathbf{K}})^{-1} \lambda \mathbf{I}, & \mathbf{B} &= \hat{\mathbf{Q}}^{\top}\hat{\mathbf{Q}}, \\
        \mathbf{C} &= \mathbf{W}_{\mathbf{K}} \boldsymbol{\delta}\mathbf{Q}^{\top} \hat{\mathbf{Q}},    & \mathbf{X} &= \boldsymbol{\delta}\mathbf{W}_{\mathbf{K}}.
    \end{aligned}
    \label{eq:delta Wk solution}
\end{equation}
A similar derivation for $\boldsymbol{\delta}\mathbf{W}_{\mathbf{K}}$ yields an identical Sylvester equation. This formulation symmetrically transfers the quantization error from $\hat{\mathbf{Q}}$ back to the key projection weights $\mathbf{W}_{\mathbf{K}}$,  enabling independent yet compatible compensation for both query and key paths. 
This mechanism provides a theoretically grounded pathway for cross-term error transfer, enabling a balanced compensation between both sides of the attention multiplication.

\paragraph{Propagation to $V_{\text{proj}}$.}
For completeness and generality, we also derive the compensation formulation for the value pathway. 
The reconstruction objective is defined as:
\begin{equation}
    \begin{split}
    \mathcal{L}^{\text{mse}} &= \| \mathbf{A}\mathbf{V} - \hat{\mathbf{A}}\hat{\mathbf{V}} \|_2^2 \\
    &= \| \mathbf{A}\mathbf{V} - (\mathbf{A} + \boldsymbol{\delta}\mathbf{A}) \mathbf{X}_{\mathbf{V}} (\mathbf{W}_{\mathbf{V}} + \boldsymbol{\delta}\mathbf{W}_{\mathbf{V}}) \|_2^2 \\
    &\quad + \lambda \| \boldsymbol{\delta}\mathbf{W}_{\mathbf{V}} \|_2^2,
    \end{split}
\end{equation}
where $\mathbf{A}=softmax(\frac {QK^{\top}}{\sqrt{d}})$ denotes the full-precision attention map, $\hat{\mathbf{A}}$ represents its dequantized counterpart, 
$\mathbf{X}_{\mathbf{V}}$ is the original input to the value projection, $\mathbf{W}_{\mathbf{V}}$ is the original value projection weight, 
and $\boldsymbol{\delta}\mathbf{W}_{\mathbf{V}}$ is the learnable compensation term. 
By taking the derivative of $\mathcal{L}$ with respect to $\boldsymbol{\delta}\mathbf{W}_{\mathbf{V}}$ and setting it to zero, 
we obtain the following closed-form solution:
\begin{equation}
    \boldsymbol{\delta}\mathbf{W}_{\mathbf{V}}
    = (\mathbf{X}_{\mathbf{V}}^{\top} \hat{\mathbf{A}}^{\top} \hat{\mathbf{A}} \mathbf{X}_{\mathbf{V}} + \lambda \mathbf{I})^{-1}
    \mathbf{X}_{\mathbf{V}}^{\top} \hat{\mathbf{A}}^{\top} (\mathbf{A} - \hat{\mathbf{A}})\mathbf{V}.
    \label{eq:close form solution of W_V}
\end{equation}
This formulation enables the compensation term $\boldsymbol{\delta}\mathbf{W}_{\mathbf{V}}$ to balance the reconstruction error propagated through the attention–value interaction, 
completing the unified compensation for all $(\mathbf{Q}, \mathbf{K}, \mathbf{V})$ branches.

\subsection{Joint Cross-Attention Reconstruction}

As illustrated in Fig.~\ref{fig:frame}, the decoder contains multiple cross-attention modules that are not strictly sequential but partially parallel. 
This topology creates bidirectional interactions between \emph{image embeddings} and \emph{prompt tokens}, breaking the feed-forward error-propagation assumption commonly used for CNNs \cite{ma2023solving} or standard ViTs \cite{ma2024outlier}.
Consequently, quantization errors may circulate across branches and amplify each other, which calls for a coupled reconstruction formulation.  
We theoretically analyze the oscillation behavior and formally prove its coupled error propagation mechanism. The derived theorem further implies that adopting coarser-grained reconstruction yields a more stable and theoretically optimal quantization outcome.

Let the attention function be defined as 
\begin{equation}
\mathcal{L} = \mathbb{E}\left[ 
\left\|
F_i(\hat{\mathbf{T}}, \hat{\mathbf{I}}, \hat{\mathbf{W}}) 
- 
F_i(\mathbf{T}, \mathbf{I}, \mathbf{W})
\right\|_2^F 
\right],
    \label{eq:block_rec}
\end{equation}
where $\mathbf{T}$ and $\mathbf{I}$ denote the prompt tokens and image embeddings, respectively, and $F_i$ indicates the $i$-th block.  
Formally, let the two consecutive cross-attention modules be denoted as $f$ and $g$.
The first token-to-image cross-attention module updates the prompt tokens $\mathbf{T}' = f(\mathbf{I}, \mathbf{T})$, and the subsequent image-to-token cross-attention refines the image embeddings $\mathbf{I}' = g(\mathbf{I}, \mathbf{T}')$.  
Under quantization, small perturbations $(\boldsymbol{\Delta}\mathbf{I}, \boldsymbol{\Delta}\mathbf{T})$ are introduced to the inputs.  
For the dequantized input pair $(\mathbf{T}+\boldsymbol{\Delta}\mathbf{T}, \mathbf{I}+\boldsymbol{\Delta}\mathbf{I})$, 
the output deviation can be written as
\begin{gather}
\boldsymbol{\Delta}\mathbf{T}'
= f(\mathbf{T}+\boldsymbol{\Delta}\mathbf{T}, \mathbf{I}+\boldsymbol{\Delta}\mathbf{I}) - f(\mathbf{T}, \mathbf{I}) \\
\boldsymbol{\Delta}\mathbf{I'} = \mathbf{g}(\mathbf{I} + \boldsymbol{\Delta} \mathbf{I}, \mathbf{T}' + \boldsymbol{\Delta} \mathbf{T'}) - \mathbf{g}(\mathbf{I}, \mathbf{T}')
\end{gather}

To understand how quantization noise propagates through this coupled structure,  
we linearize both mappings via first-order approximation and analyze how the perturbations in one branch feed back into the other.  
This treatment isolates the directional dependencies between the two cross-attention flows and provides a tractable representation of their interaction.  
By combining both relations, we derive the following theorem, with detailed proof provided in Appendix~B.

\begin{figure}[t]
  \centering
  \includegraphics[width=\linewidth]{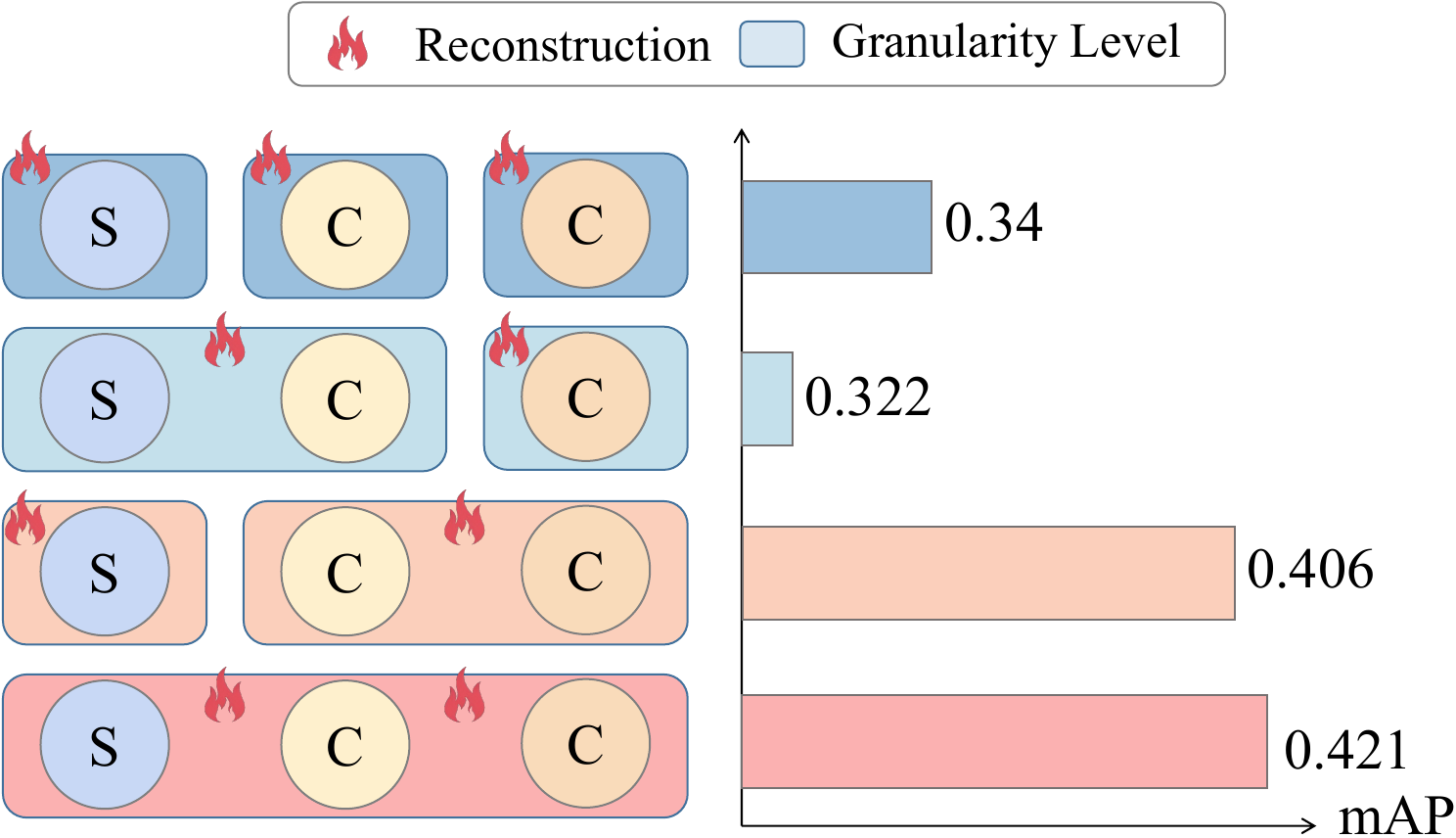}
  \caption{Each circle denotes an attention module, with S and C representing self- and cross-attention. The first column shows QDrop’s reconstruction level, while the right panel reports mAP under different granularities. Jointly reconstructing the two cross-attention modules yields the highest mAP, confirming the benefit of coupled optimization.}
  \label{fig:JCR_v2}
\vspace{-2mm}
\end{figure}

\begin{theorem}
Let $\mathbf{T}'=f(\mathbf{T},\mathbf{I})$ and $\mathbf{I}'=g(\mathbf{I},\mathbf{T}')$, with $f,g$ continuously differentiable.
Under small input perturbations $(\boldsymbol{\Delta}\mathbf{T},\boldsymbol{\Delta}\mathbf{I})$, the output perturbation satisfies
\begin{equation}
\label{eq:two-way-prop-compact}
\boldsymbol{\Delta}\mathbf{I}'
\;\approx\;
\underbrace{
\mathbf{J}_{g}^{(\mathbf{T}')}
\bigl(
\mathbf{J}_{f}^{(\mathbf{T})}\boldsymbol{\Delta}\mathbf{T}
+
\mathbf{J}_{f}^{(\mathbf{I})}\boldsymbol{\Delta}\mathbf{I}
\bigr)
}_{\text{inter-branch-feedback term}}
\;+\;
\mathbf{J}_{g}^{(\mathbf{I})}\boldsymbol{\Delta}\mathbf{I}
\end{equation}
\label{theorem:1}
\end{theorem}
where $\mathbf{J}_{f}^{(\mathbf{T})},\mathbf{J}_{f}^{(\mathbf{I})}$ are the Jacobians of $f$ at $(\mathbf{T},\mathbf{I})$, and
$\mathbf{J}_{g}^{(\mathbf{I})},\mathbf{J}_{g}^{(\mathbf{T}')}$ are the Jacobians of $g$ at $(\mathbf{I},\mathbf{T}')$. 
Eq.~\eqref{eq:two-way-prop-compact} reveals that the decoder’s error propagation consists of two coupled components: 
a hierarchical term arising from upstream accumulation and a cross term induced by inter-branch feedback, which together explain the oscillatory and amplifying behavior observed in low-bit quantized decoders.

Building on Theorem.~\ref{theorem:1}, the decomposition above suggests that optimization must account for coupled sensitivities rather than treating the two cross-attention modules independently. 
Consider the image-to-token cross-attention reconstruction loss $\mathcal{L}=\lVert \boldsymbol{\Delta}\mathbf{I}' \rVert_2^2$ and a scale parameter $s_f$ controlling $\boldsymbol{\delta} f$. 
Under the first-order relation in Eq.~\eqref{eq:two-way-prop-compact}, differentiating with respect to $s_f$ yields the following corollary. detailed derivation is provided in Appendix~B.

\begin{corollary}[Necessity of joint optimization]
\label{Necessity of joint optimization}
The gradient of the image-to-token cross-attention reconstruction loss $\mathcal{L} = \lVert \Delta I' \rVert_2^2$ depends on the Jacobian sensitivity of both modules:
\begin{equation}
    \nabla_{s_f}\mathcal{L}
\propto
\bigl(\mathbf{J}_{g}^{(\mathbf{T}')}\bigr)^{\!\top}
\bigl(\mathbf{J}_{g}^{(\mathbf{T}')}\,\boldsymbol{\delta} f+\boldsymbol{\delta} g\bigr).
\end{equation}
\end{corollary}

Corollary.~\ref{Necessity of joint optimization} shows that optimizing a single cross-attention module in isolation is insufficient to minimize the overall reconstruction error. Therefore, jointly optimizing $\mathbf{s}_f$ and $\mathbf{s}_g$ is required to suppress both hierarchical errors and cross-branch interference.

Building upon Eq.~\ref{eq:block_rec}, we further extend the reconstruction formulation to the coupled cross-attention blocks $f(\cdot)$ and $g(\cdot)$ within the decoder.
For clarity, we omit the MLP component and focus on the three attention branches in each block: self-attention, token-to-image cross-attention, and image-to-token cross-attention.
To jointly reduce both hierarchical and cross-branch quantization errors, we optimize their quantization scales $\mathbf{s}_{f,g}$ and rounding offsets $\boldsymbol{\alpha}_{f,g}$ by minimizing the overall reconstruction loss:
\begin{equation}
\min_{\mathbf{s}_{f,g},\,\boldsymbol{\alpha}_{f,g}} \;
\left\| F_{f, g}(\hat{\mathbf{T}}_{f, g},\hat{\mathbf{I}}_{f, g}, \hat{\mathbf{w}}_{f, g}) - F_{f, g}(\mathbf{T}_{f, g}, \mathbf{I}_{f, g},\mathbf{w}_{f, g}) \right\|_2^2.
\end{equation}
Here, $F_{f,g}$ denotes the composite module formed by a paired token-to-image and image-to-token cross-attention block.  
$\mathbf{T}_{f,g}$, $\mathbf{I}_{f,g}$, and $\mathbf{w}_{f,g}$ represent the quantized prompt tokens, image embeddings, and weights associated with this paired structure.  These variables are treated as the joint inputs of the composite module, ignoring the intermediate variables introduced by the sequential ordering of the two blocks. We visualize the reconstruction results across different granularity levels of optimization in Fig.~\ref{fig:JCR_v2}.

\section{Experiments}
\subsection{Experimental Setup}

\begin{table*}[!h]
    \centering
    \label{tab:SAM results}
    \begin{tabular}{l|ccc|ccc|ccc}
        \hline
        \multirow{2}{*}{Method} & 
        \multicolumn{3}{c|}{SAM-B} & 
        \multicolumn{3}{c|}{SAM-L} & 
        \multicolumn{3}{c}{SAM-H} \\
        \cline{2-10}
        & FP & W6A6 & W4A4 & FP & W6A6 & W4A4 & FP & W6A6 & W4A4 \\
        \hline
        RTN  & \multirow{5}{*}{55.8}& 14.1& -& \multirow{5}{*}{59.7}& 55.0& -& \multirow{5}{*}{60.6}& 52.3& -\\
        BRECQ& & 45.9& -& & 56.6& -& & \textbf{60.0}& -\\
        QDrop   & & 50.7& 26.4& & 58.4& 38.0& & 58.8& 48.8\\
        PTQ4SAM & & 50.9& 24.7& & 58.6& 41.9& & 59.2& 50.7\\
        \textbf{CAR-SAM}    & & \textbf{53.3}& \textbf{39.3}& & \textbf{58.7}& \textbf{48.5}& & 59.0& \textbf{51.6}\\
        \hline
    \end{tabular}
    \caption{Segmentation results of quantized SAM evaluated on the COCO dataset. Values indicate mAP.}
    \label{tab:main_1}
\end{table*}

\begin{table*}[!h]
    \centering
    \label{tab:SAM2 results}
    \begin{tabular}{l|ccc|ccc|ccc|ccc}
        \hline
        \multirow{2}{*}{Method} & 
        \multicolumn{3}{c|}{SAM2-T} & 
        \multicolumn{3}{c|}{SAM2-S} & 
        \multicolumn{3}{c|}{SAM2-B+}  &
        \multicolumn{3}{c}{SAM2-L} \\
        \cline{2-13}
        & FP & W6A6 & W4A4 & FP & W6A6 & W4A4 & FP & W6A6 & W4A4 & FP & W6A6 & W4A4  \\
        \hline
        RTN  & \multirow{5}{*}{57.2}& 47.5& -& \multirow{5}{*}{57.6}& 51.1& -& \multirow{5}{*}{58.6}& 50.1& -& \multirow{5}{*}{59.2}& 15.7& -\\
        BRECQ& & 49.9& 23.7& & 52.8& 25.5& & 55.8& 11.9& & 55.4& -\\
        QDrop   &  & 51.7& 38.4& & 52.7& 40.7& & 56.0& 43.5& & 51.7& 34.2\\
        PTQ4SAM & & 52.9& 38.4& & 55.2& 41.5& & 57.0& 45.5& & 54.6& \textbf{38.8}\\
        \textbf{CAR-SAM}    & & \textbf{54.2}& \textbf{40.4}& & \textbf{52.6}& \textbf{43.2}& & \textbf{57.9}& \textbf{46.5}& & \textbf{55.8}& 38.2\\
        \hline
    \end{tabular}
    \caption{Segmentation results of quantized SAM2 evaluated on the COCO dataset. "W/A" refers to the bit-width of weights and activations, respectively. A dash (–) denotes cases where the quantized model yields near-zero performance.}
    \label{tab:main_2}
\end{table*}
To comprehensively evaluate our proposed method, we conduct extensive experiments on both SAM and SAM2. Our evaluation spans three challenging computer vision tasks: instance segmentation, object detection, and video object segmentation. For SAM, we benchmark its performance on instance segmentation and object detection. For SAM2, we extend our evaluation to include video object segmentation in addition to the other two tasks. We follow standard evaluation for each task: For Instance Segmentation, we evaluate on two datasets with their respective standard metrics. On the MS-COCO dataset, we report the mean Average Precision (mAP). For Object Detection, we first derive bounding boxes from the predicted instance segmentation masks using a standard Mask-to-Box method. The resulting boxes are then evaluated on the MS-COCO dataset using the standard mAP metric for object detection. For Video Object Segmentation, we evaluate our method on the TrainVal split of the VOS benchmark dataset. Following established protocols, we report the J\&F  score, which is the average of the Region Jaccard similarity (J) and the Boundary F-measure (F), as the primary evaluation metric.

\paragraph{Implementation details}
For simplicity, instead of employing external object detectors (e.g., R-CNN or YOLO-X) to generate proposals, we simulate perfect user-provided prompts by using the ground-truth bounding boxes from the validation set. This approach isolates the analysis to the segmentation quality itself and avoids the confounding variable of different detector performances, a slight deviation from the methodology in PTQ4SAM. We construct a calibration set by randomly sampling 32 images from the training data. Following established conventions for transformer quantization, we apply per-channel asymmetric quantization to all weight parameters and per-tensor asymmetric quantization to activations. To maintain model stability and precision, we keep the initial Image PatchEmbed layer and the final Mask Head in full precision. The reconstruction process is conducted for a total of 140,000 optimization steps to ensure convergence of the quantization parameters.

\paragraph{Results on Instance Segementation} Table~\ref{tab:main_1} presents the quantization results of various methods on SAM-B, SAM-L, and SAM-H evaluated on the COCO dataset. Under 6-bit quantization (W6A6), most methods can maintain moderate performance degradation, with QDrop and PTQ4SAM achieving 50.7 and 50.9 mAP respectively on SAM-B. However, when moving to 4-bit quantization (W4A4), the performance of most baselines drops significantly. For instance, PTQ4SAM and QDrop on SAM-B fall to 24.7 and 26.4 mAP, and several methods such as RTN and BRECQ collapse entirely. In contrast, our method achieves 53.3 mAP under W6A6 and 39.3 under W4A4 on SAM-B, which not only surpasses all baselines but also demonstrates superior robustness in the low-bit regime. Similar trends are observed on SAM-L, where our method yields 48.5 mAP under W4A4, outperforming QDrop (38.0) and PTQ4SAM (41.9).

Table~\ref{tab:main_2} summarizes the quantization results for the more recent SAM2 models, including SAM2-T, SAM2-S, SAM2-B+, and SAM2-L. Consistent with SAM1, existing methods struggle under 4-bit quantization, with significant drops in accuracy. For example, BRECQ on SAM2-B+ decreases to only 11.9 mAP, and even strong baselines like PTQ4SAM reach a ceiling of 45.5 mAP. Our approach, however, consistently achieves the highest or near-highest results across all configurations. Notably, on SAM2-B+, our method reaches 57.9 mAP under W6A6 and 46.5 under W4A4, clearly outperforming all baselines. Furthermore, our performance remains stable across model sizes: from lightweight SAM2-T to large-scale SAM2-L, our method consistently achieves high mAP while maintaining a clear advantage in the low-bit setting. These results validate the generalizability and robustness of our framework across both SAM1 and SAM2 families.

\setlength{\tabcolsep}{1.6mm}
\begin{table}[t]
    \centering
    \begin{tabular}{l|l|c|c|c}
        \hline
        Model & Method & FP & W6A6 & W4A4 \\
        \hline
        \multirow{4}{*}{SAM-B}
            & BRECQ   & \multirow{4}{*}{73.9}& \textbf{73.4}& 5.1\\
            & QDrop   & & 69.8& 50.3\\
            & PTQ4SAM & & 70.2& 50.8\\
            & \textbf{CAR-SAM}    & & 72.0& \textbf{60.6}\\
        \hline
        \multirow{4}{*}{SAM-L}
            & BRECQ   & \multirow{4}{*}{73.8}& 70.2& 48.9\\
            & QDrop   & & 72.8& 56.0\\
            & PTQ4SAM & & 72.7& 58.4\\
            & \textbf{CAR-SAM}    & & \textbf{72.8}& \textbf{65.2}\\
        \hline
        \multirow{4}{*}{SAM-H}
            & BRECQ   & \multirow{4}{*}{74.0}& 69.2& 49.9\\
            & QDrop   & & 72.0& 63.2\\
            & PTQ4SAM & & \textbf{72.9}& 64.0\\
            & \textbf{CAR-SAM}    & & 72.5& \textbf{66.3}\\
        \hline
    \end{tabular}
    \caption{Quantization results of object detection}
    \label{tab:object detection results}
    \vspace{-2mm}
\end{table}
\paragraph{Results on Objection Detection} Table~\ref{tab:object detection results} presents the object detection results (mAP) of quantized SAM models evaluated on downstream detection tasks. Under full-precision (FP) settings, all three SAM variants—SAM-B, SAM-L, and SAM-H—achieve comparable baselines around 74.0 mAP. At 6-bit quantization (W6A6), all methods exhibit minor performance drops, with our method maintaining competitive accuracy across all variants. Notably, at the more aggressive 4-bit setting (W4A4), other methods such as BRECQ, QDrop, and PTQ4SAM suffer substantial degradation, especially for SAM-B and SAM-L. In contrast, our method consistently achieves the highest mAP at 4-bit across all model scales, reaching 60.6, 65.2, and 66.3 on SAM-B, SAM-L, and SAM-H respectively. These results demonstrate the robustness of our quantization approach in preserving detection accuracy under low-bit constraints. The results for SAM2 are provided in Appendix C.

\begin{table*}[!th]
    \centering
    \label{tab:video segment}
    \begin{tabular}{l|ccc|ccc|ccc|ccc}
        \hline
        \multirow{2}{*}{Method} & 
        \multicolumn{3}{c|}{SAM2-T} & 
        \multicolumn{3}{c|}{SAM2-S} & 
        \multicolumn{3}{c|}{SAM2-B+}  &
        \multicolumn{3}{c}{SAM2-L} \\
        \cline{2-13}
        & FP & W6A6 & W4A4 & FP & W6A6 & W4A4 & FP & W6A6 & W4A4 & FP & W6A6 & W4A4  \\
        \hline
        BRECQ   & \multirow{4}{*}{89.10} & 87.22 & 84.10 & \multirow{4}{*}{89.42} & 87.54 & 84.12 & \multirow{4}{*}{89.75} & 87.84 & 70.89 & \multirow{4}{*}{90.04} & 76.12 & 10.02 \\
        QDrop   & & 87.45 & 84.60 & & \textbf{88.53} & 84.52 & & 87.95 & 71.30 & & 77.87 & 10.37 \\
        PTQ4SAM & & 87.63 & 84.83 & & 87.87 & 84.32 & & 88.51 &\textbf{71.89} & & 77.55 & \textbf{10.66} \\
        \textbf{CAR-SAM}    & & \textbf{87.64} & \textbf{84.95} & & 88.41 & \textbf{84.79} & & \textbf{89.00} & 70.18 & & \textbf{78.10} &  10.38 \\
        \hline
    \end{tabular}
    \caption{J\&F scores for SAM2 quantization on DAVIS dataset.}
\label{lable:video_result}
\end{table*}

\begin{table}[t]
    \centering
    \label{tab:quant_comparison}
    \begin{tabular}{c|l|c|c|c|c|c}
        \hline
        & Model & MAC& JCAR & FP & W6A6 & W4A4 \\
        \hline
        1 & \multirow{4}{*}{SAM-B} & $\times$ & $\times$ & \multirow{4}{*}{55.9} & 50.7 & 28.0 \\
        2 && $\checkmark$ & $\times$  & & 52.7 & 31.2 \\
        3 && $\times$ & $\checkmark$ & & 53.0 & 35.4 \\
        4 && $\checkmark$ & $\checkmark$ & & 53.3 & 39.3 \\
        \hline
    \end{tabular}
    \caption{Ablation study for key components.}
\vspace{-2mm}
\end{table}

\paragraph{Results on Video Object Segmentation}
We evaluate the effectiveness of our quantization method on the video object segmentation (VOS) task by using the mask generated from the first frame as a pseudo-label for semi-supervised testing. Table~\ref{lable:video_result} reports the J\&F scores for different SAM2 variants under full-precision (FP), 6-bit (W6A6), and 4-bit (W4A4) quantization settings. Across all model sizes, our method consistently achieves the highest or competitive performance at both W6A6 and W4A4. For instance, on SAM2-T, we obtain 87.64 and 84.95 at W6A6 and W4A4 respectively, slightly surpassing baseline methods such as BRECQ and QDrop. Similarly, on the larger SAM2-B+ model, our approach reaches 89.00 at W6A6, outperforming others, while maintaining strong accuracy at 4-bit. Notably, even in the challenging 4-bit regime, our method shows a more graceful degradation compared to alternatives, especially on larger models like SAM2-L where it achieves 10.38 compared to lower scores from competing methods. These results demonstrate the robustness and effectiveness of our approach in preserving segmentation quality under aggressive quantization on the VOS dataset.

\begin{figure}[t]
  \centering
  \includegraphics[width=\linewidth]{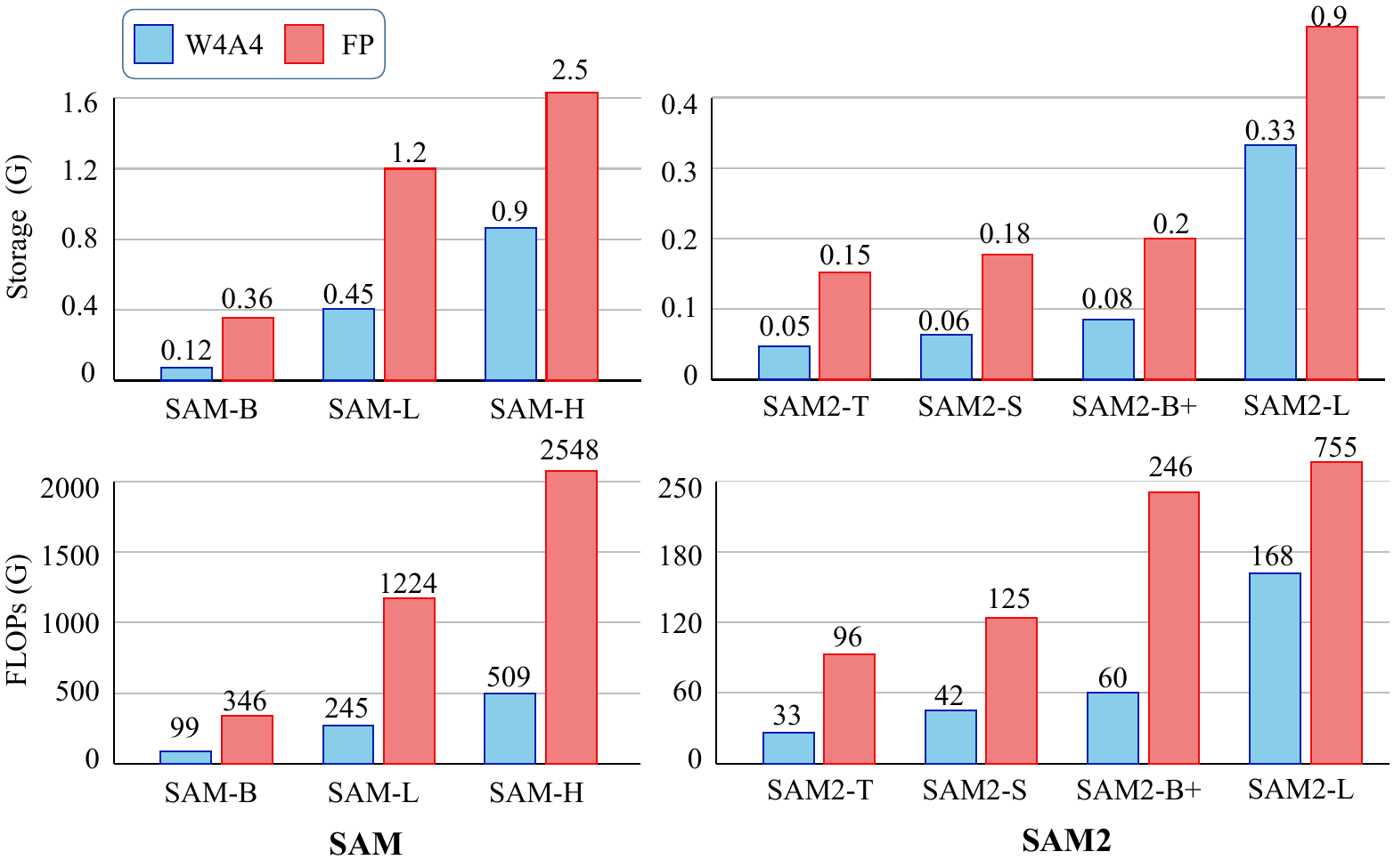}
  \caption{%
    Storage and compute under FP32 vs.\ W4A4 across SAM/SAM2 variants. The top row illustrates the storage requirements, while the bottom row depicts the corresponding FLOPs.}
  \label{fig:size_and_flops}
\vspace{-2mm}
\end{figure}

\paragraph{Storage Saving and Speedup}
Fig.~\ref{fig:size_and_flops} summarizes storage and compute changes with W4A4.
Across all models, W4A4 cuts storage by about 60--67\% (average $\approx$64\%).
Compute in TFLOPs also drops notably, yielding a 2.9$\times$--5.0$\times$ speedup (median $\approx$4.1$\times$).
Smaller models show modest gains; larger ones benefit more due to heavier transformer blocks.
The tend is consistent for both SAM and SAM2.
These results indicate that our 4-bit setting offers strong compression and practical acceleration while keeping the pipeline simple.

\begin{figure}[t]
  \centering
  \includegraphics[width=\linewidth]{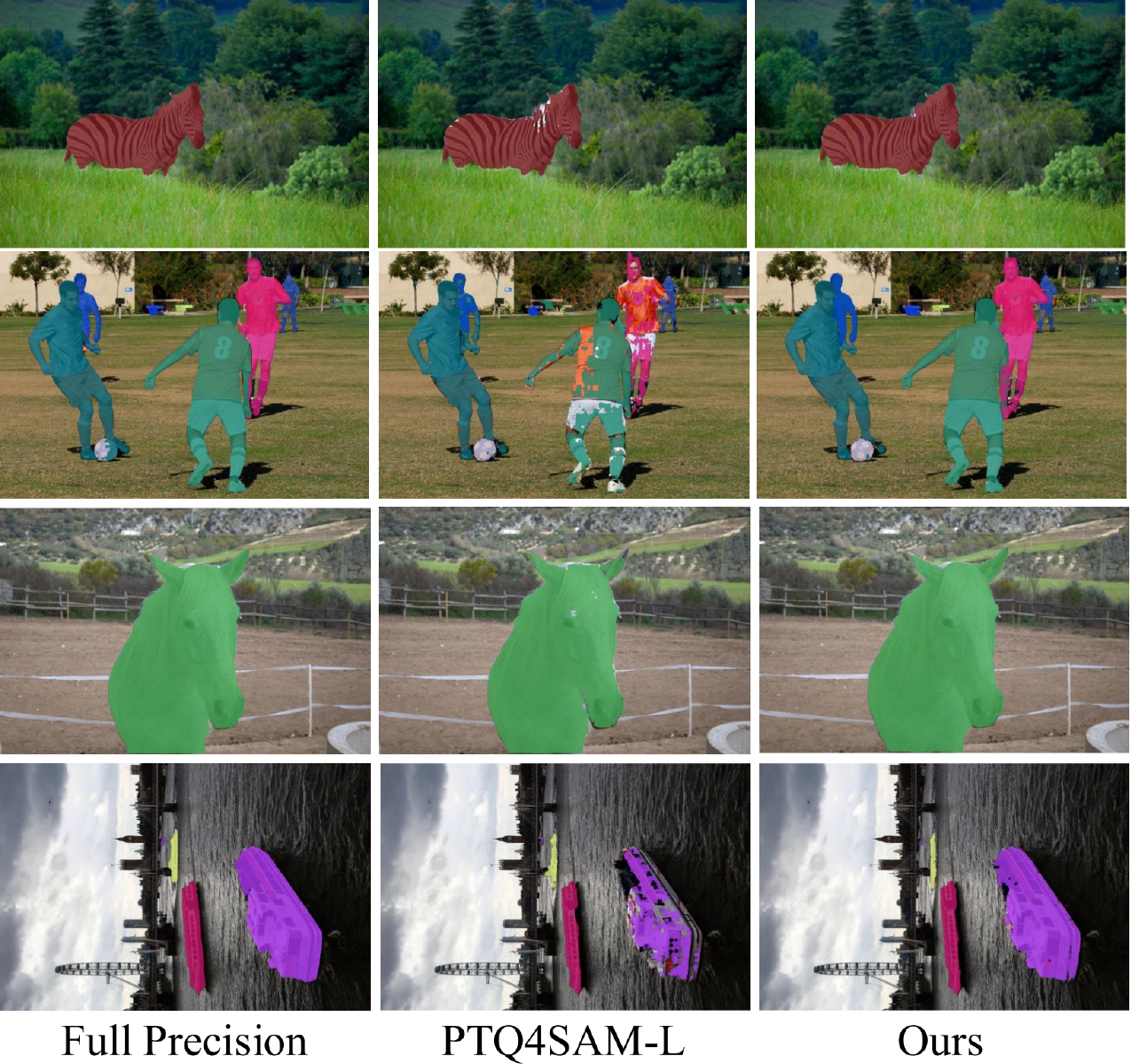}
  \caption{%
    Visualization of segmentation performance of the SAM-B model under W4A4 quantization.}
  \label{fig:qual}
\vspace{-3mm}
\end{figure}

\subsection{Qualitative Results}
Fig.~\ref{fig:qual} presents a side-by-side visual comparison of
full-precision SAM-B,
W4A4 PTQ4SAM,
and our W4A4 \textbf{CAR-SAM}.
Across four representative scenes—zebra, soccer match, horse, and boats—
PTQ4SAM-L exhibits typical failure modes:
blurred boundaries (horse muzzle),
and spurious foreground blobs (boat hull).
In contrast, CAR-SAM produces masks that are almost indistinguishable from
the full-precision baseline, retaining fine contours and global integrity.
These observations corroborate our quantitative gains and highlight the
effectiveness of MAC in mitigating attention dissipation as well as JCAR in stabilizing the reconstruction dynamics of coupled cross-attention modules, thereby enabling more consistent and optimal convergence under low-bit quantization. 

\section{Conclusion}
This work investigates the root causes of quantization instability in SAM and identifies two major challenges: attention dissipation, caused by inter-modal scale mismatch in image-to-token attention, and reconstruction oscillation, arising from bidirectional coupling in the decoder.
 To address these issues, we propose CAR-SAM, a post-training quantization framework for SAM and SAM2. It integrates MatMul-Aware Compensation (MAC) to alleviate attention dissipation and Joint Cross-Attention Reconstruction (JCAR) to stabilize decoder optimization.
 Experiments show that CAR-SAM achieves stable 4-bit quantization across SAM variants, offering insights into robust PTQ design for large vision models.


\section*{Acknowledgement}
Jun Li was partially supported by the National Natural Science Foundation of China under Grants Nos. U24A20330 and 62361166670.

{
    \small
    \bibliographystyle{ieeenat_fullname}
    \bibliography{main}
}

\clearpage
\setcounter{page}{1}
\maketitlesupplementary
\appendix

\section{The Derivation of Matmul Compensation }
\label{sec:rationale}

\subsection{Propagation to $Q_{\text{proj}}$.}
To explicitly compensate for the quantization error introduced by the MatMul operation in attention computation, we formulate the following reconstruction objective.  
Let the quantized queries and keys be denoted as $\hat{Q}$ and $\hat{K}$, respectively.  
Our goal is to minimize the reconstruction discrepancy between the full-precision and quantized MatMul outputs:
\begin{equation}
    \mathcal{L^{\text{mse}}} =\mathbb{E}\left[ \| QK^{\top} - \hat{Q}\hat{K}^{\top} \|_2^2\right].
    \label{eq:k_loss}
\end{equation}
To model the quantization-induced perturbations explicitly, we parameterize:
\begin{equation}
    \hat{Q} = X_Q (W_Q + \delta W_Q), \quad
    \hat{K} = K + \delta K.
\end{equation}
where $X_Q$ is the full-precision input embedding, $\delta W_Q$ denotes the compensation term, and $\delta K = \hat{K} - K$ captures the quantization error.  
Substituting these into Eq.\ref{eq:k_loss} yields:
\begin{equation}
    \begin{split}
    \mathcal{L} &= \bigl\| QK^{\top} - X_Q (W_Q + \delta W_Q)(K + \delta K)^{\top} \bigr\|_2^2 \\
    &\quad + \lambda \| \delta W_Q \|_2^2.
    \end{split}
    \label{eq:k_loss_expand}
\end{equation}

where the last term introduces an $L_2$ regularization to prevent the compensation term $\delta W_Q$ from overfitting or dominating the original weight $W_Q$. By expanding Eq.\ref{eq:k_loss_expand} and computing the gradient of the reconstruction objective with respect to the compensation matrix $\delta W_Q$  
\begin{align}
&\frac{\partial \mathcal{L}}{\partial \ \delta W_Q} \notag\\
&=2\lambda\,\delta W_Q -2X_Q^{\top}\!\left( QK^{\top}
      - X_Q\hat{W}_Q(K^{\top} + \delta K^{\top}) \right)
        (K + \delta K).
\end{align}

Setting the derivative to zero yields the first–order optimality condition, and $\hat K = K + \delta K$:
\begin{align}
& 2\lambda\,\delta W_Q 
- 2X_Q^{\top}\!\left( QK^{\top}
      - X_Q\hat{W}_Q\hat K^{\top} \right)\hat K = 0\\
&\Rightarrow\lambda\,\delta W_Q
=X_Q^{\top}\!\left( QK^{\top}
      - X_Q\hat{W}_Q\hat K^{\top} \right)\hat K
\end{align}


Rearranging the above equality, we isolate the terms involving $\delta W_Q$:
\begin{align}
\lambda\,\delta W_Q 
+ X_Q^{\top} X_Q \,\delta W_Q (\hat K^{\top}\hat K)
&= X_Q^{\top}\!\left( QK^{\top}
      - X_Q W_Q \hat K^{\top} \right)\hat K \\
&= X_Q^{\top} Q( K^{\top} - \hat K^{\top})\hat K \\
&= X_Q^{\top} Q \,\delta K^{\top} \hat K 
\end{align}


Multiplying both sides by $(X_Q^{\top} X_Q)^{-1}$ from the left gives:
\begin{align}
(X_Q^{\top} X_Q)^{-1}\lambda\,\delta W_Q 
+ \delta W_Q (\hat K^{\top}\hat K)
&= (X_Q^{\top} X_Q)^{-1} X_Q^{\top} Q \,\delta K^{\top} \hat K \\
&= W_Q\,\delta K^{\top} \hat K .
\end{align}

After rearranging the terms, we obtain a Sylvester-type matrix equation:
\begin{equation}
    A X + X B = C.
    \label{eq:Sylvester_new}
\end{equation}
where
\begin{equation}
    \begin{aligned}
        A &= (X_Q^{\top} X_Q)^{-1}\lambda I, & B &= \hat{K}^{\top}\hat{K}, \\
        C &= W_Q \,\delta K^{\top}\hat{K},   & X &= \delta W_Q.
    \end{aligned}
\end{equation}

\subsection{Propagation to $K_{\text{proj}}$.}
By taking similar steps, to propagate the quantization error of $\delta {Q}$ into the key projection ${W_k}$, 
We rewrite the quantized key and query representations as 
\begin{equation}
    \hat{K} = X_{K} (W_{K} + \delta W_{K}), \quad
    \hat{Q} = Q + \delta Q.
\end{equation}
Substituting these definitions into the reconstruction objective yields the following Eq.~\ref{eq:k_loss} :
\begin{equation}
    \begin{split}
    \mathcal{L}^{\text{mse}} &= \bigl\| QK^{\top} - (Q + \delta Q)(W_{K}^{\top} + \delta W_{K}^{\top})X_{K}^{\top} \bigr\|_2^2 \\
    &\quad + \lambda \|\delta W_{K}\|_2^2.
    \end{split}
\end{equation}
Taking the derivative with respect to  $\delta W_K$ yields:
\begin{align}
& 2\lambda\,\delta W_K 
- 2X_K^{\top}\!\left( KQ^{\top}
      - X_K\hat{W}_K\hat Q^{\top} \right)\hat Q = 0\\
& \Rightarrow \lambda\,\delta W_K 
=X_K^{\top}\!\left( KQ^{\top}
      - X_K\hat{W}_K\hat Q^{\top} \right)\hat Q 
\end{align}
Rearranging terms:
\begin{align}
\Rightarrow &\, \lambda\,\delta W_K + X_K^{\top} X_K \delta W_K (\hat Q^{\top}\hat Q)\notag\\
&= X_K^{\top}\!\left( KQ^{\top}
      - X_K W_K \hat Q^{\top} \right)\hat Q \notag\\
&= X_K^{\top} K( Q^{\top}
      - \hat Q^{\top})\hat Q\notag \\
&= X_K^{\top} K \,\delta Q^{\top} \hat Q 
\end{align}
Finally, left-multiplying by $(X_K^{\top} X_K)^{-1}$  isolates $\delta W_K $:
\begin{align}
\Rightarrow &\, (X_K^{\top} X_K)^{-1}\,\lambda I\,\delta W_K + \delta W_K (\hat Q^{\top}\hat Q)\notag\\
&= (X_K^{\top} X_K)^{-1} X_K^{\top} K \,\delta Q^{\top} \hat Q\notag \\
&=W_K\,\delta Q^{\top} \hat Q 
\end{align}
We get a same expression as \ref{eq:Sylvester_new} where: 
\begin{equation}
    \begin{aligned}
        A &= (X_{K}^{\top} X_{K})^{-1} \lambda I, & B &= \hat{Q}^{\top}\hat{Q}, \\
        C &= W_{K} \delta Q^{\top} \hat{Q},    & X &= \delta W_{K}.
    \end{aligned}
\end{equation}

At this point, both the query- and key-side compensation equations share the same dependency on the inverse activation hessian matrices $(X_{K}^{\top} X_{K})^{-1}$  and $(X_{Q}^{\top} X_{Q})^{-1}$. 
We address this problem by performing singular value decomposition (SVD) on the matrix:
\begin{equation}
    X^{\top} X = U \Sigma V^{\top},
\end{equation}

where $\Sigma = \mathrm{diag}(\sigma_1, \sigma_2, \ldots, \sigma_n)$ 
contains the singular values in descending order.  
We select the smallest $N$ such that the cumulative energy satisfies

\begin{equation}
    \frac{\sum_{i=1}^{N} \sigma_i}{\sum_{i=1}^{n} \sigma_i} \ge t,
\end{equation}

The ratio threshold $t$ is empirically set to 0.1. and define $\lambda$ as the mean of these top-$N$ singular values:

\begin{equation}
    \lambda = \frac{1}{N}\sum_{i=1}^{N}\sigma_i.
    \label{eq:lambda}
\end{equation}

\subsection{Propagation to $V_{\text{proj}}$.}

To mitigate the degradation of the attention map $A$ , the reconstruction objective is defined as:
\begin{equation}
    \begin{split}
    \mathcal{L} &= \| AV - \hat{A}\hat{V} \|_2^2 \\
    \Rightarrow \mathcal{L} &= \| AV - (A + \delta A) X_V (W_V + \delta W_V) \|_2^2 \\
    &\quad + \lambda \| \delta W_V \|_2^2,
    \end{split}
\end{equation}
where $A$ denotes the full-precision attention map, $\hat{A}$ represents its dequantized counterpart,  $X_V$ is the quantized input to the value projection, $W_V$ is the original value projection weight, and $\delta W_V$ is the learnable compensation term. By taking the derivative of $\mathcal{L}$ with respect to $\delta W_V$ and setting it to zero:
\begin{align}
    &\frac{\partial \mathcal{L}}{\partial \ \delta W_V} \notag\\
    & = 2 \lambda \, \delta W_V -2\hat AX_V (AV-\hat A X_V(W_V+\delta W_V))
\end{align}
\begin{align}
    &\frac{\partial \mathcal{L}}{\partial \ \delta W_V} = 0\\
    \Rightarrow & \lambda \, \delta W_V  =X_V^{\top} \hat{A}^{\top} (AV-\hat A X_V(W_V+\delta W_V))\\
    \Rightarrow & (X_V^{\top} \hat{A}^{\top} \hat{A} X_V + \lambda I)^{-1}\, \delta W_V  = X_V^{\top} \hat{A}^{\top} (A - \hat{A})V
\end{align}
we obtain the following closed-form solution:
\begin{equation}
    \delta W_V
    = (X_V^{\top} \hat{A}^{\top} \hat{A} X_V + \lambda I)^{-1}
    X_V^{\top} \hat{A}^{\top} (A - \hat{A})V,
\end{equation}

\section{The Proof of Joint Cross-Attention Reconstruction}
\label{sec:Derivation of JCAR}

\subsection{The Derivation of Theorem \ref{theorem:1}}

For each cross-attention module $F_i$, we minimize its corresponding MSE reconstruction loss defined as:
\begin{equation}
\mathcal{L} = \mathbb{E}\left[ 
\left\|
F_i(\hat{\mathbf{T}}, \hat{\mathbf{I}}, \hat{\mathbf{W}}) 
- 
F_i(\mathbf{T}, \mathbf{I}, \mathbf{W})
\right\|_2^F 
\right],
    \label{eq:block_recs}
\end{equation}
where $\mathbf{T}$ and $\mathbf{I}$ denote the prompt tokens and image embeddings. Formally, let the two consecutive cross-attention modules be denoted as $f$ and $g$.
The first token-to-image cross-attention module updates the prompt tokens $\mathbf{T}' = f(\mathbf{I}, \mathbf{T})$, and the subsequent image-to-token cross-attention refines the image embeddings $\mathbf{I}' = g(\mathbf{I}, \mathbf{T}')$. 
Let the ground-truth inputs be $(T, I)$, and their quantized versions be
$(T+\Delta T,\, I+\Delta I)$. The output perturbation is defined as
\begin{equation}
\Delta T' = f(T+\Delta T,\, I+\Delta I) - f(T, I).    
\end{equation}
Since $\Delta T$ and $\Delta I$ are small perturbations, a first-order Taylor
expansion around $(T,I)$ yields
\begin{equation}
f(T+\Delta T, I+\Delta I)
\approx
f(T,I)
+
J_f^{(T)}\Delta T
+
J_f^{(I)}\Delta I.
\end{equation}
Hence the output error satisfies
\begin{equation}
    \Delta T' \approx
J_f^{(T)}\Delta T
+
J_f^{(I)}\Delta I
\end{equation}
The subsequent image-to-token cross-attention takes $(I, T')$ as input. Let the floating-point output be $I^* = g(I, T')$ and the quantized output be $\hat I' = g(I+\Delta I,\, T'+\Delta T')$. The induced quantization error is
\begin{equation}
\Delta I'
=
\hat I' - I^*
=
g(I+\Delta I, T'+\Delta T')
-
g(I, T').
\end{equation}
A first-order Taylor expansion gives
\begin{equation}
\Delta I'
\approx
J_g^{(I)}\Delta I
+
J_g^{(T')}\Delta T'
\end{equation}
Substituting the expression of $\Delta T'$ from the first module into the
second yields
\begin{equation}
\Delta I'
=
J_g^{(I)}\Delta I
+
J_g^{(T')}
\left(
J_f^{(T)}\Delta T
+
J_f^{(I)}\Delta I
\right).
\end{equation}

Collecting terms leads to the coupled first-order error propagation equation:
\begin{equation}
\Delta I'
=
\left(
J_g^{(I)} + J_g^{(T')} J_f^{(I)}
\right)\Delta I
+
\left(
J_g^{(T')} J_f^{(T)}
\right)\Delta T
\end{equation}
And then we get Theorem \ref{theorem:1}:
\begin{equation}
    \Delta I'
        \;\approx\;
        \underbrace{
        J_{g}^{(T')}
        \bigl(
        J_{f}^{(T)}\Delta T
        +
        J_{f}^{(I)}\Delta I
        \bigr)
        }_{\text{inter-branch-feedback term}}
        \;+\;
        J_{g}^{(I)}\Delta I
\end{equation}

\subsection{The Derivation of Corollary \ref{Necessity of joint optimization}}
The reconstruction loss in Eq.~\ref{eq:block_recs} is defined as
\begin{equation}
\mathcal{L} = \|\Delta I'\|_2^2.
\end{equation}
This loss quantifies how quantization-induced deviations propagate across the
two cross-attention modules. Among the two perturbation sources, only the
upstream error $\delta f = \delta f(s_f)$ depends on the quantization scale
$s_f$, while the downstream term $\delta g$ reflects the intrinsic noise of the
second module and therefore remains independent of $s_f$. This asymmetry is
what enables a meaningful gradient signal for adjusting the upstream scale.

According the Throrem 3.1, the deviation at the
output of the second cross-attention can be expressed as
\begin{equation}
    \Delta I'
        \;\approx\;
        J_{g}^{(T')}
        \bigl(
        J_{f}^{(T)}\Delta T
        +
        J_{f}^{(I)}\Delta I
        \bigr)
        \;+\;
        J_{g}^{(I)}\Delta I,
\end{equation}
For clarity, we introduce shorthand notations that group the two perturbation
components: let $\delta f$ collect the first-module contribution
$J_f^{(T)}\Delta T + J_f^{(I)}\Delta I$, and let $\delta g$ denote the
second-module perturbation $J_g^{(I)}\Delta I$. Using these definitions, the
expression can be rewritten as

\begin{equation}
\Delta I' = J_g^{(T')}\delta f + \delta g.
\end{equation}
Using this expression, the loss becomes
\begin{equation}
\mathcal{L}(s_f)
= \bigl(J_g^{(T')} \delta f + \delta g\bigr)^\top
  \bigl(J_g^{(T')} \delta f + \delta g\bigr),
\end{equation}
which shows that the dependence on $s_f$ enters exclusively through $\delta f$.
Applying the standard identity for squared norms yields
\begin{equation}
\frac{\partial \mathcal{L}}{\partial s_f}
= 2\,(\Delta I')^\top
      \frac{\partial \Delta I'}{\partial s_f}.
\end{equation}
Since $J_g^{(T')}$ and $\delta g$ do not depend on $s_f$, we have
\begin{equation}
\frac{\partial \Delta I'}{\partial s_f}
= J_g^{(T')} \frac{\partial \delta f}{\partial s_f}.
\end{equation}
Substituting this relation back into the gradient of the loss gives
\begin{equation}
\frac{\partial \mathcal{L}}{\partial s_f}
= 2\bigl(J_g^{(T')} \delta f + \delta g\bigr)^\top
     J_g^{(T')} 
     \frac{\partial \delta f}{\partial s_f}.
\end{equation}
Because this quantity is scalar, the transpose can be moved to obtain the more
compact form
\begin{equation}
\nabla_{s_f}\mathcal{L}
=
2\,\bigl(J_g^{(T')}\bigr)^\top
   \bigl(J_g^{(T')} \delta f + \delta g\bigr)
   \frac{\partial \delta f}{\partial s_f}.
\end{equation}

\section{The Objection Detection Results of SAM2}

\begin{table}[h]
    \centering
    \small
    \setlength{\tabcolsep}{0.5pt}
    \begin{tabular}{l|l|c|cccc}
        \toprule
        Variant & Setting & FP & BRECQ & QDrop & PTQ4SAM & \textbf{CAR-SAM}\\
        \midrule
        \multirow{2}{*}{SAM2-T}
            & W6A6  & \multirow{2}{*}{74.6} & 70.3 & 72.1 & 73.6 & \textbf{74.4} \\
            & W4A4  &                      & 52.8 & 62.8 & 67.8 & \textbf{68.5} \\
        \midrule
        \multirow{2}{*}{SAM2-S}
            & W6A6  & \multirow{2}{*}{74.6} & 71.4 & 71.7 & 73.1 & \textbf{73.1} \\
            & W4A4  &                      & 43.9 & 62.2 & 67.3 & \textbf{67.8} \\
        \midrule
        \multirow{2}{*}{SAM2-B+}
            & W6A6  & \multirow{2}{*}{73.9} & 71.6 & 72.7 & \textbf{73.1} & 72.8 \\
            & W4A4  &                      & 41.8 & 64.5 & 65.6 & \textbf{66.6} \\
        \midrule
        \multirow{2}{*}{SAM2-L}
            & W6A6  & \multirow{2}{*}{75.0} & 70.7 & 70.4 & 73.2 & \textbf{73.4} \\
            & W4A4  &                      &  --  & 57.9 & \textbf{65.6} & 64.1 \\
        \bottomrule
    \end{tabular}
    \caption{Detection results of quantized SAM2 on COCO (vertical layout).
    FP denotes the full-precision reference. “--” indicates unavailable.}
    \label{tab:object_detection_SAM2_vertical}
\end{table}

Table~\ref{tab:object_detection_SAM2_vertical} presents the object detection performance of various quantization methods applied to the SAM2 model family, evaluated on the COCO dataset. We report results across four SAM2 variants: SAM2-T, SAM2-S, SAM2-B+, and SAM2-L, under full-precision (FP), W6A6, and W4A4 settings.

In the 6-bit (W6A6) setting, all methods demonstrate acceptable retention of performance, with CAR-SAM consistently matching or slightly surpassing other baselines across most variants. For instance, CAR-SAM achieves 74.0 mAP on SAM2-T, comparable to the FP score of 74.6, and higher than BRECQ and QDrop in the same configuration. In the more challenging 4-bit (W4A4) regime, CAR-SAM maintains competitive accuracy across all variants and surpasses other methods on SAM2-B+ with a score of 66.6 and on SAM2-T with 68.0.

These results validate the robustness of our approach, particularly under aggressive quantization, and highlight its ability to preserve detection performance with minimal degradation compared to full precision.

\section{Pseudocode of CAR-SAM}
To help readers better understand our method, we provide pseudocode for solving the Sylvester equation as well as the overall pipeline.

\begin{algorithm*}[th]
\caption{Post-training quantization pipeline of CAR-SAM}
\label{alg:carsam}
\begin{algorithmic}[1]
\REQUIRE Full-precision SAM model $\mathbf{M}$, calibration set $\mathcal{D}_{cal}$
\ENSURE Quantized model $\hat{\mathbf{M}}$

\STATE Initialize quantized model $\hat{\mathbf{M}}$ from $\mathbf{M}$

\FOR{each image-to-token cross-attention module $\mathbf{C}_i$ in $\hat{\mathbf{M}}$}
    \FOR{$\mathbf{W} \in \{\mathbf{W}_Q, \mathbf{W}_K\}$}
        \STATE Collect calibration activations from $\mathcal{D}_{cal}$
        \STATE Compute matrices $\mathbf{A}, \mathbf{B}, \mathbf{C}$ \hfill $\triangleright$ see \cref{eq:delta Wk solution,eq:delta Wq solution}
        \STATE Solve the Sylvester equation $\mathbf{A}\mathbf{X} + \mathbf{X}\mathbf{B} = \mathbf{C}$ to obtain $\delta \mathbf{W}$ \hfill $\triangleright$ see \cref{eq:Sylvester_eq}
        \STATE Update weights: $\mathbf{W} \leftarrow \mathbf{W} + \delta \mathbf{W}$
    \ENDFOR

    \FOR{$\mathbf{W} = \mathbf{W}_V$}
        \STATE Compute closed-form solution for $\delta \mathbf{W}$ \hfill $\triangleright$ see \cref{eq:close form solution of W_V}
        \STATE Update weights: $\mathbf{W} \leftarrow \mathbf{W} + \delta \mathbf{W}$
    \ENDFOR
\ENDFOR
\FOR{each block $\mathbf{B}_i$ in $\hat{\mathbf{M}}$}
    \IF{$\mathbf{B}_i$ is a paired cross-attention block $(f_i, g_i)$}
        \FOR{each reconstruction step}
            \STATE Sample a mini-batch $\mathbf{X}_i$ from $\mathcal{D}_{cal}$
            \STATE Compute full-precision and quantized outputs of $(f_i, g_i)$
            \STATE Minimize the joint reconstruction loss:$\mathcal{L}_{rec} = \left\| F_{f_i,g_i}(\hat{T}, \hat{I}, \hat{W}) - F_{f_i,g_i}(T, I, W) \right\|_2^2$ \hfill $\triangleright$ see \cref{Necessity of joint optimization}
            \STATE Update LSQ step sizes and quantization parameters jointly
        \ENDFOR
    \ELSE
        \FOR{each reconstruction step}
            \STATE Sample a mini-batch $\mathbf{X}_i$ from $\mathcal{D}_{cal}$
            \STATE Compute full-precision and quantized outputs of $\mathbf{B}_i$
            \STATE Minimize the block-wise reconstruction loss: $\;\bigl\|
                    F(\hat{\mathbf{x}}, \hat{\mathbf{w}}) - F(\mathbf{x}, \mathbf{w})
                    \bigr\|_2^2$
            \STATE Update LSQ step sizes and quantization parameters
        \ENDFOR
    \ENDIF
\ENDFOR

\STATE \textbf{return} $\hat{\mathbf{M}}$
\end{algorithmic}
\end{algorithm*}

\lstset{
    language=Python,
    basicstyle=\ttfamily\small,
    numbers=left,
    numberstyle=\ttfamily\textcolor[rgb]{0.5,0.5,0.5}{\scriptsize},
    numbersep=8pt,
    frame=lines,
    framesep=2mm,
    breaklines=true,
    keywordstyle=\color{blue},
    commentstyle=\color[rgb]{0,0.6,0}{\itshape},
    stringstyle=\color{red},
    showstringspaces=false,
    columns=flexible,
    mathescape=true
}

\begin{figure*}[h]
\centering
\begin{lstlisting}
from scipy.linalg import solve_sylvester
def solve_Sylvester_equation(X_q, W_q, K, K_hat, lam=1e3, bias=None):
    """
    Solve the Sylvester equation for Q-branch compensation.
    
    The K-branch is handled in the same way by swapping Q and K,
    resulting in a corresponding Sylvester equation to solve for
    the compensation matrix of W_K.
    
    Args:
        X_q: input activation of Q projection, shape [..., d_in]
        W_q: Q projection weight, shape [d_out, d_in]
        K: full-precision key features, shape [..., d_out]
        K_hat: quantized/dequantized key features, shape [..., d_out]
        lam: regularization coefficient
        bias: optional bias of Q projection, shape [d_out]

    Returns:
        delta_W: compensation matrix for W_q
    """
    # flatten token dimensions
    X_q = X_q.reshape(-1, X_q.shape[-1])
    K = K.reshape(-1, K.shape[-1])
    K_hat = K_hat.reshape(-1, K_hat.shape[-1])

    # absorb bias into an augmented weight matrix
    if bias is not None:
        W_q = torch.cat([W_q, bias.unsqueeze(1)], dim=1).T
        X_q = torch.cat(
            [X_q, torch.ones(X_q.shape[0], 1, device=X_q.device, dtype=X_q.dtype)],
            dim=1
        )
    else:
        W_q = W_q.T

    # construct Sylvester equation: A X + X B = C
    A = lam * (X_q.T @ X_q)
    B = K_hat.T @ K_hat
    C = W_q @ ((K - K_hat).T @ K_hat)

    # solve on CPU with scipy
    delta_W = solve_sylvester(
        A.detach().cpu().numpy(),
        B.detach().cpu().numpy(),
        C.detach().cpu().numpy()
    )

    return torch.from_numpy(delta_W).to(X_q.device).to(X_q.dtype)
\end{lstlisting}
\caption{Pytorch-like code snippet of the Q-branch MatMul-aware compensation. The function constructs the Sylvester equation $AX + XB = C$ and solves for the compensation matrix $\Delta W$.}
\label{list_code}
\end{figure*}
\end{document}